\documentclass[a4paper]{article}

\usepackage{INTERSPEECH2022}
\usepackage{float}
\usepackage{arydshln}

\title{Joint Encoder-Decoder Self-Supervised Pre-training for ASR}
\name{A Arunkumar, S Umesh}
\address{
  Speech Lab, Indian Institute of Technology Madras}
\email{arunkumaras10@gmail.com, umeshs@ee.iitm.ac.in}

\begin{document}

\maketitle
\begin{abstract}
  Self-supervised learning (SSL) has shown tremendous success in various speech-related downstream tasks, including Automatic Speech Recognition (ASR). The output embeddings of the SSL model are treated as powerful short-time representations of the speech signal. However, in the ASR task, the main objective is to get the correct sequence of acoustic units, characters, or byte-pair encodings (BPEs). Usually, encoder-decoder architecture works exceptionally well for a sequence-to-sequence task like ASR. Therefore, in this paper, we propose a new paradigm that exploits the power of a decoder during self-supervised learning. We use Hidden Unit BERT (HuBERT) SSL framework to compute the conventional masked prediction loss for the encoder. In addition, we have introduced a decoder in the SSL framework and proposed a target preparation strategy for the decoder. Finally, we use a multitask SSL setup wherein we jointly optimize both the encoder and decoder losses. We hypothesize that the presence of a decoder in the SSL model helps it learn an acoustic unit-based language model, which might improve the performance of an ASR downstream task. We compare our proposed SSL model with HuBERT and show up to 25\% relative improvement in performance on ASR by finetuning on various LibriSpeech subsets.
\end{abstract}

\section{Introduction}
\label{intro}
Self-Supervised Learning (SSL) techniques \cite{oord2018representation, chung2019unsupervised, schneider2019wav2vec, baevski2019vq, liu2020mockingjay, baevski2020wav2vec, 9053569, 9478264, hsu2021hubert, chen2021wavlm} utilize large volumes of unlabeled speech data to learn high-level representations that work well for various speech-related downstream tasks. These representations have also been shown to work well for Automatic Speech Recognition (ASR) task where the objective is to get a sequence of acoustic units or their manifestation in terms of characters or Byte Pair Encodings (BPEs) \cite{sennrich2015neural}. 
There are several approaches in self-supervised learning in speech domain including those based on autoregressive predictive coding \cite{chung2019unsupervised}, contrastive losses \cite{oord2018representation,schneider2019wav2vec,baevski2019vq, baevski2020wav2vec}, masked prediction \cite{liu2020mockingjay, 9478264, hsu2021hubert, chen2021wavlm} and multi-task learning \cite{9053569}. Most SSL models are encoder-only models. In Hidden Unit BERT (HuBERT) \cite{hsu2021hubert}, acoustic units are discovered using clustering techniques, and these units are used as targets in computing the masked prediction loss at the encoder output. 

In supervised ASR tasks, encoder-decoder architectures \cite{lu2015study,8462506} are very popular since the decoder brings in the advantage of learning the implicit language model and outputting the desired sequence of acoustic units or characters. To bring the same advantage to SSL, this work proposes to incorporate a decoder in the SSL framework. The additional advantage is that the decoder can now better capture the sequential characteristics of the discovered acoustic units.

The main contributions of this paper are as follows: 
\begin{itemize}
    \item A transformer decoder \cite{vaswani2017attention} is introduced in the SSL framework in addition to the encoder, and we name this SSL technique as Joint Encoder-Decoder Pre-training. 
    \item An unsupervised technique to generate the target sequence for the decoder is proposed
    \item Self-supervised learning is now done with the combined objective of masked prediction loss similar to conventional HuBERT and the newly introduced sequence loss of the decoder.
    \item The downstream ASR model is obtained by finetuning the proposed Encoder-Decoder SSL Model with downstream supervised data using joint CTC/Attention loss \cite{7953075} in a multitask learning framework. 
\end{itemize}

The rest of the paper is organized as follows. Section \ref{proposed} briefly explains the working of existing HuBERT and the architectural changes resulting in the proposed joint encoder-decoder pre-training method. The target sequence preparation for the decoder is also discussed in detail in the same section. Section \ref{experiments} describes the ASR finetuning setup. Sections \ref{scratchresults}, \ref{cpresults} and \ref{randomdecresults} present the results of various experiments conducted. Section \ref{conclusions} draws important conclusions from this work and discusses future work. Pre-trained HuBERT-Base model used in the experiments was downloaded from fairseq \cite{ott2019fairseq}. The proposed SSL model was implemented using ESPnet toolkit \cite{watanabe2018espnet}, and all the experiments were conducted using the same.

\section{Proposed Joint Encoder-Decoder Pre-training Method}
\label{proposed}
The proposed Joint Encoder-Decoder SSL method introduces a decoder in addition to the encoder in the conventional SSL framework. We have considered HuBERT SSL in this work. The architecture of the conventional HuBERT SSL and the additional blocks that make up the proposed Joint Encoder-Decoder SSL are explained in this section. 
\subsection{Conventional HuBERT SSL}
\label{hubert}
\begin{figure}[H]
    \centering
    \includegraphics[scale=0.6]{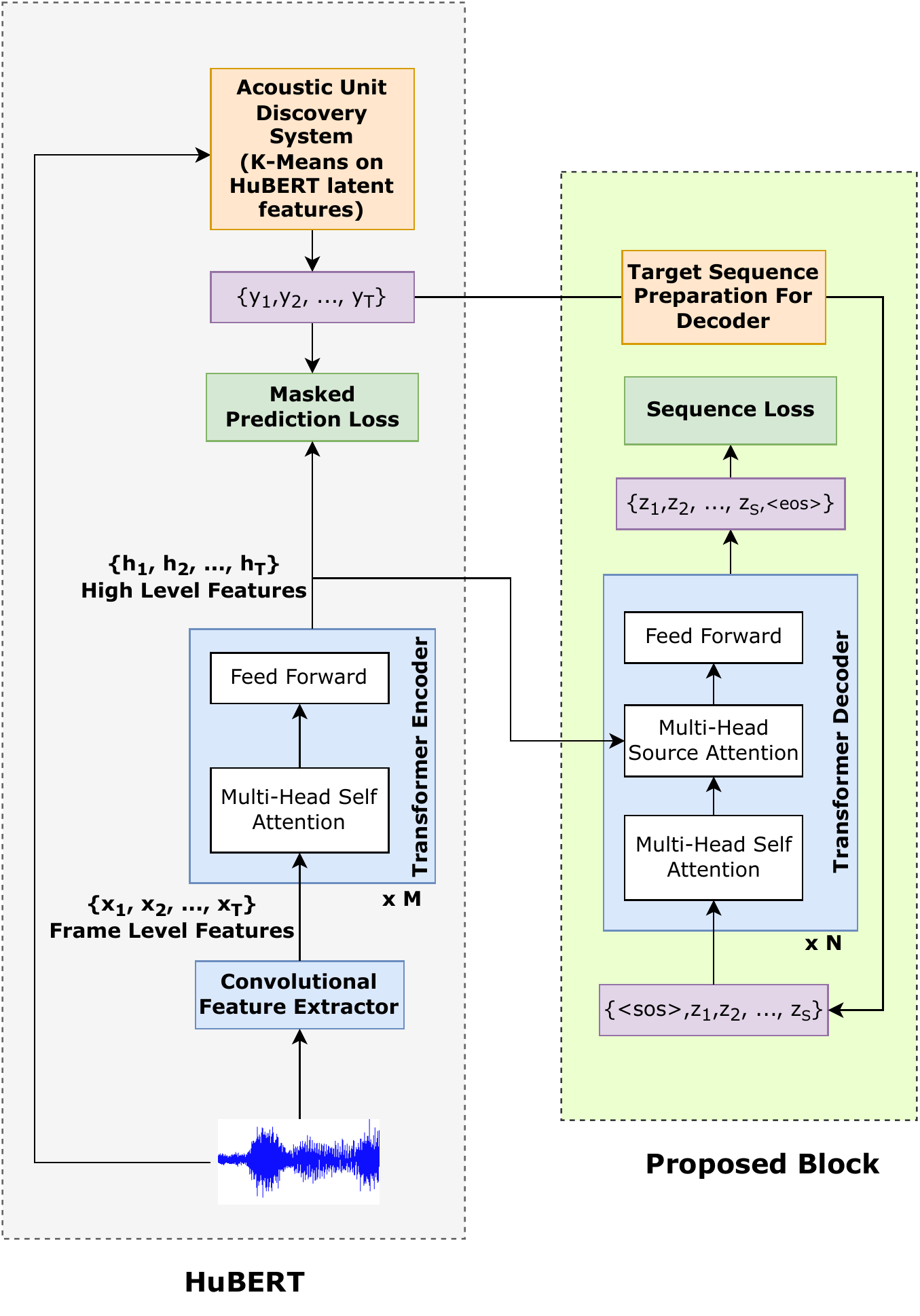}
    \caption{Proposed Architecture}
    \label{fig:arch}
\end{figure}

 HuBERT is a non-contrastive SSL model trained on raw speech waveforms, and its architecture is shown in the left block of Figure \ref{fig:arch}. A convolutional feature extractor transforms the raw speech waveforms into the frame-level features. The feature extractor is frozen for the downstream tasks. A transformer encoder block receives masked feature frames according to a masking logic.
The masking logic assigns a selection probability to each frame. Ten consecutive frames are masked for every selected frame, including the selected frame. The masked frames are fed to a Transformer Encoder which transforms them through several encoder layers, and each encoder output frame is then fed to a classification layer. Targets for the classification layer are the discovered cluster IDs of the corresponding input frames. The cluster ID corresponding to a frame is obtained by building a K-means clustering model with the MFCC features or HuBERT intermediate layer features extracted for the training data. A frame is assigned to a cluster if its distance to that cluster center is minimal. In this work, we always use the sixth layer features extracted from the existing pre-trained HuBERT-Base model for training the K-means model. Based on the weights assigned to the cluster ID prediction of unmasked frames, the contribution of unmasked frames to the final loss is decided. The prediction loss for masked frames alone was shown to be sufficient in \cite{hsu2021hubert}, and therefore, we also use only masked prediction loss.
\\

\subsection{Proposed Joint Encoder-Decoder SSL} 
\subsubsection{Architecture}
\label{arch}
The proposed SSL method makes architectural changes to the conventional HuBERT SSL model by incorporating a Transformer decoder, as shown in Figure \ref{fig:arch}. Decoders have the ability to model and generate sequences. In ASR, the decoder acts as an implicit language model, and Transformer encoder-decoder ASR models are very popular. It might be beneficial to incorporate the Transformer decoder during SSL too. The added SSL decoder is made to attend to the SSL encoder outputs through source attention like in any other Transformer based encoder-decoder model. 
\subsubsection{Decoder Target Preparation}
\label{targetprep}
\begin{figure}[H]
    \centering
    \includegraphics[scale=0.65]{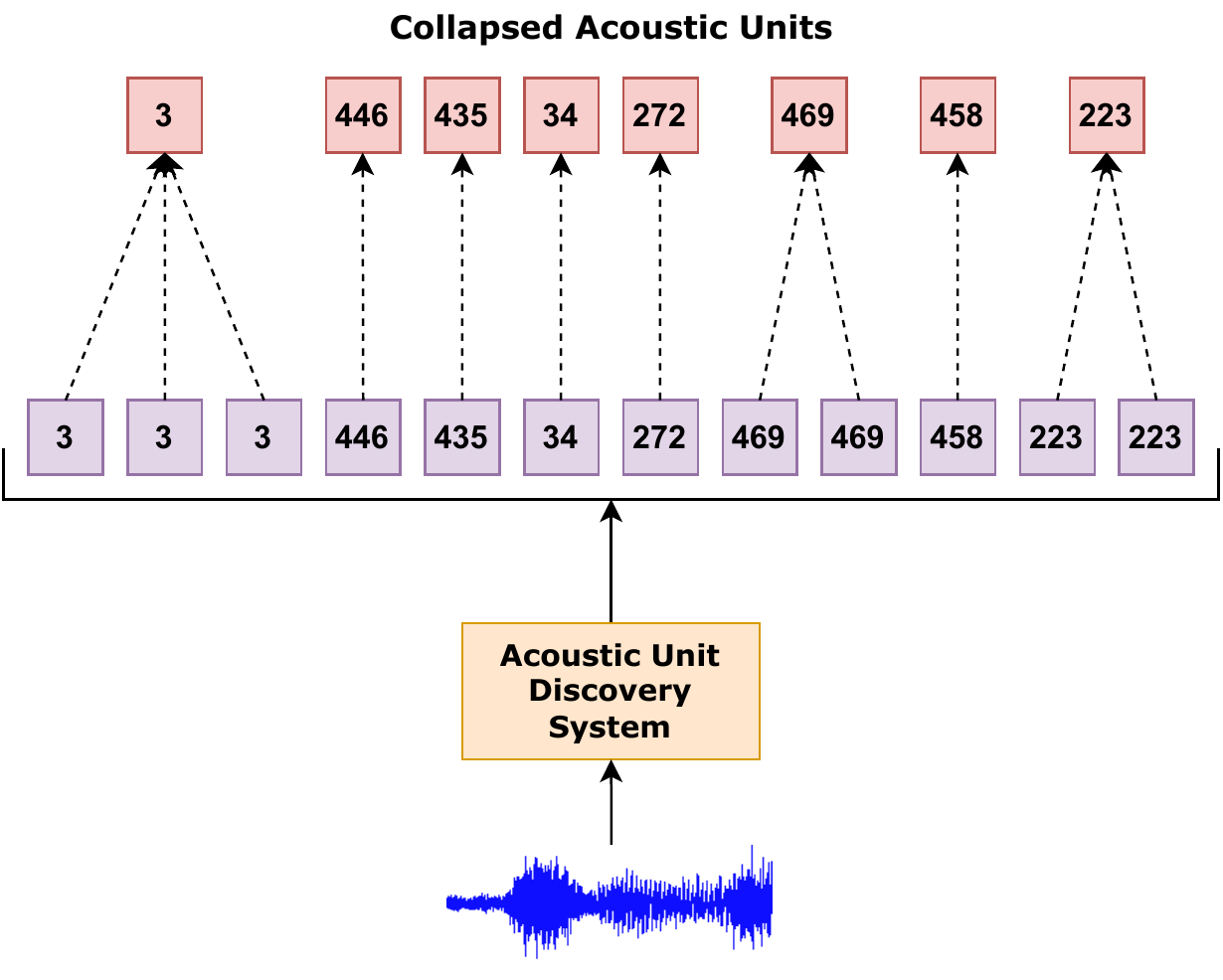}
    \caption{Target Sequence Preparation}
    \label{fig:seq}
\end{figure}

Figure \ref{fig:seq} shows the approach that we have adopted to obtain in an unsupervised manner the targets for training the added SSL Transformer decoder. Targets for the added decoder are prepared from the discovered target cluster IDs of all the encoder input frames, both masked and unmasked. Any consecutive repetition in cluster IDs is replaced by a single cluster ID. Hence the decoder targets are also obtained by self-supervision. The core idea behind collapsing repeated target cluster IDs of the encoder frames for the decoder target preparation is to make the decoder learn the sequential characteristics of the acoustic units, and this might be beneficial to the downstream Transformer based Encoder-Decoder ASR.

\subsubsection{Loss Function for the Proposed SSL Model}
\label{loss}
Let $L_M$ be the masked prediction loss from the encoder and $L_S$ be the sequence loss from the decoder. Then the total loss, $L$ will be a weighted sum of both the losses as given by,
 
 \begin{equation}
   L = \alpha * L_M + (1 - \alpha) * L_S  
 \end{equation}
Equal importance to both $L_M$ and $L_S$ is given by setting 
$\alpha$ to 0.5. So, while pre-training, the encoder weights are affected by both the losses during backpropogation.

\section{Finetuning the SSL Models for ASR}
\label{experiments}
The SSL models are usually evaluated on various downstream tasks by finetuning the SSL models with the task-specific loss functions. In this work, the ASR task is considered. The experimental setups for the ASR task with the conventional and proposed SSL technique are presented in the following subsections.

\subsection{Finetuning the HuBERT SSL Model for ASR}
\label{baselinefinetuning}
For performing the ASR task with HuBERT, it is finetuned by using labeled data from different subsets of Libri-light. 
A randomly initialized CTC layer is added on top of the HuBERT final layer and finetuned to predict the characters. 
The loss function is taken as plain CTC. This ASR model serves as the baseline for evaluating the merit of the proposed SSL technique.

\subsection{Finetuning the Proposed SSL Model for ASR}
\label{proposedfinetuning}
The proposed SSL model has both encoder and decoder, as mentioned in Section \ref{arch}. For performing ASR in this case, both the encoder and decoder of the proposed SSL model are finetuned. The classification layers on top of the encoder and decoder are randomly initialized as done in HuBERT finetuning and finetuned to predict character outputs. The loss function for training this ASR model is the joint CTC/attention loss given by,
\begin{equation}
L = \beta * CTC\ Loss + (1 - \beta) * Attention\ Loss
\end{equation}

The CTC loss is contributed by the encoder, and attention loss is contributed by the decoder. $\beta$ is set to 0.3. The final loss is a weighted sum of both these losses. This is similar to the standard ASR recipe used in espnet.



\section{Results of ASR Finetuning of SSL Models Trained from Scratch}
\label{scratchresults}
LibriSpeech-360h \cite{panayotov2015librispeech} was used to train the proposed SSL model from scratch for 101 epochs. LibriSpeech-960h was not used owing to compute resource constraints. For a fair comparison, the baseline HuBERT SSL model was also trained from scratch with LibriSpeech-360h since a 360-hour HuBERT pre-trained model is not available in the public domain. Both the baseline and proposed SSL model have 12 encoder layers with 8 attention heads and 3072 linear units. The number of decoder layers in the proposed SSL model is 8, with 8 attention heads and 2048 linear units. The SSL models were trained with 15 Million batch-bins and a learning rate of 0.0001. Warmup scheduler with 25000 warmup steps and Adam optimizer were used. 


\begin{table}[H]
\centering
\caption{Results of ASR Finetuning of SSL Models Trained from Scratch and without Language Model (LM)}
\label{tab:results_scratch}
\begin{tabular}{@{}ccc@{}}
\toprule
Model           & test  & test  \\
   &clean &other\\
\hline \hline
\multicolumn{3}{l}{\textbf{1h labeled}}     \\
HuBERT-360h (ours)          & 47.4        & 62.0        \\
Proposed SSL-360h  & \textbf{39.4}        & \textbf{56.2}        \\
\hline \hline
\multicolumn{3}{l}{\textbf{10h labeled}}    \\
HuBERT-360h (ours)          & 29.5        & 47.4        \\
Proposed SSL-360h  & \textbf{22.6}        & \textbf{41.0}        \\
\hline \hline
\multicolumn{3}{l}{\textbf{100h labeled}}   \\
HuBERT-360h (ours)          & 15.5        & 36.0        \\
Proposed SSL-360h  & \textbf{11.5}        & \textbf{30.6}        \\ 
\bottomrule
\end{tabular}
\end{table}
For ASR finetuning, 3.2 Million batch-bins with a learning rate of 0.00002 were used. Warmup scheduler with 8000 warmup steps and Adam optimizer were used. The results of the downstream ASR tasks obtained by finetuning the respective SSL models are presented in Table \ref{tab:results_scratch}. We have used three subsets of Libri-light to compare our proposed method with HuBERT baseline.

From Table~1, the following observations can be made:
\begin{itemize}
    \item ASR finetuning of the proposed SSL model is superior to the baseline HuBERT SSL for all the Libri-light subsets taken.
    \item The relative improvements obtained from the proposed SSL increase with the increase in the amount of labeled data available with a maximum relative improvement of 25.8 \% for the LibriSpeech-100h labeled subset.
\end{itemize}

\section{Results of ASR Finetuning of SSL Models Trained by Continued Pre-training}
\label{cpresults}
The previous section described the experiments where the SSL models were trained from scratch. Due to resource constraints, LibriSpeech-360h was used instead of the bigger subset, LibriSpeech-960h. To show that our proposed model performs better than the baseline HuBERT SSL in a larger data setting, too, we opted for a continued pre-training approach on the readily available HuBERT-Base SSL model pre-trained with LibriSpeech-960h. 
\\
To get the equivalent of the proposed Joint Encoder-Decoder SSL model in the continued pre-training setup, a randomly initialized decoder was added to the already pre-trained 960-hour HuBERT-Base encoder. The pre-training was then \emph{continued} in the same manner as mentioned in the proposed SSL method. The pre-training was continued for 40 epochs.

\begin{table}[H]
\centering
\caption{Results of ASR Finetuning of SSL Models Trained by Continued Pre-training and without LM}
\label{tab:results_cp}
\begin{tabular}{@{}ccc@{}}
\toprule
Model           & test  & test  \\ 
   &clean &other \\
\hline \hline

\multicolumn{3}{l}{\textbf{1h labeled}}     \\
HuBERT-Base (Cont. Pre. Train)         & 32.0        & 40.6        \\
\hline
Proposed SSL with  & \textbf{24.0}        & \textbf{31.2}        \\
HuBERT-Base Enc. + Rand. Init. Dec. \\
 (Cont. Pre. Train) \\

\hline \hline
\multicolumn{3}{l}{\textbf{10h labeled}}    \\
HuBERT-Base (Cont. Pre. Train)          & 12.0        & 19.4        \\
\hline
Proposed SSL with & \textbf{9.8}        & \textbf{16.4}        \\
HuBERT-Base Enc.+ Rand. Init. Dec.
\\(Cont. Pre. Train)
\\
\hline \hline
\multicolumn{3}{l}{\textbf{100h labeled}}   \\
HuBERT-Base (Cont. Pre. Train)       & 6.7        & 15.1        \\
\hline
 Proposed SSL with &  \textbf{5.8}       & \textbf{13.3}         \\
(HuBERT-Base Enc. + Rand. Init. Dec.)\\
(Cont. Pre. Train)\\
\bottomrule
\multicolumn{3}{p{7.0cm}}{\textbf{Note}: Enc. denotes Encoder, Rand. Init. Dec. denotes Randomly Initialized Decoder, Cont. Pre. Train denotes Continuosly Pre-trained}
\end{tabular}
\end{table}

For the baseline, instead of directly finetuning the HuBERT-Base model for the CTC-based ASR task, we chose to continue the pre-training of HuBERT-Base for 40 epochs to ensure fairness. This is because the pre-trained HuBERT-Base model would have used a different mapping for the acoustic units than the one used in our models. For example, an acoustic unit represented by cluster ID 100 in the pre-trained HuBERT-Base model may be represented by cluster ID 25 in our model.
\\
 The finetuned ASR results for the above mentioned SSL models with continued pre-training are presented in Table \ref{tab:results_cp}. From the Table \ref{tab:results_cp}, it can be seen that
 \begin{itemize}
     \item Proposed SSL outperforms the conventional HuBERT base model in large data setting too.
     \item The results are better than the SSL models trained from scratch, owing to the bigger LibriSpeech-960h data.
     \item In this case, the encoder from base HuBERT model is already well-trained and then the randomly initialized decoder is added for continued pre-training. When both encoder and decoder are jointly optimized while training from scratch, the relative improvement over conventional HuBERT is even better as seen in Section~4.
 \end{itemize}

\section{Results of ASR Finetuning with Randomly Initialized Decoder in the SSL Models}
\label{randomdecresults}
This section presents experiments that reinforce the idea that the added decoder in the proposed SSL holds significant importance in the downstream ASR. The decoder and encoder are jointly pre-trained with combined masked prediction and sequence loss in a self-supervised way. This architecture works best for downstream ASR tasks. It is important to note that the improvements due to the proposed model are not just because of the decoder being finetuned during finetuning for the downstream ASR task. For example, we could take the pre-trained HuBERT-360 hour model and add a randomly initialized decoder to it and do a finetuning step with joint CTC/attention losses. However, as shown in Table~3, this will give worse results than HuBERT-360 because the encoder is well pre-trained while the decoder and cross-attention have to be learned with only finetuning data. A similar observation can be made when we randomized the decoder before finetuning our proposed SSL model. Again the performance degrades when compared to our original SSL model. This proves that pre-training the decoder is really helpful and improves downstream ASR performance. The results show that pre-training the decoder is helpful and improves downstream ASR performance.
\\
The corresponding results from Table~1 are reproduced in this table for comparison with the proposed model.
\begin{table}[H]
\centering
\caption{Results of ASR Finetuning with Randomly Initialized Decoder in the SSL Models and without LM for LibriSpeech-10h}
\label{tab:results_scratch_random_decoder}
\begin{tabular}{@{}ccc@{}}
\toprule
Model           & test  & test  \\
 &clean &other \\
\hline \hline
Hubert-360 (ours)     &29.5            &47.4\\
\hline
Proposed SSL-360h    &\textbf{22.6}  &\textbf{41.0}\\
\hline
HuBERT-360h (ours)        & 32.0            & 42.0            \\
+ Rand. Init. Dec. \\
\hline
Enc. of the Proposed SSL-360h   & 27.8            & 45.5   \\
+ Rand. Init. Dec.
\\ \bottomrule
\multicolumn{3}{p{7.0cm}}{\textbf{Note}: Enc. denotes Encoder, Rand. Init. Dec. denotes Randomly Initialized Decoder}
\end{tabular}
\end{table}
\section{Conclusions and Future Work}
\label{conclusions}

In this paper, a Joint Encoder-Decoder Self-Supervised Pre-training model that jointly optimizes masked prediction loss from encoder and sequence loss from decoder was proposed in place of conventional encoder-only SSL techniques. Through various experiments, the proposed SSL model was proved to be superior to the conventional encoder-only HuBERT SSL model for the downstream ASR task. In the future, we plan to evaluate the proposed model by pre-training with LibriSpeech 960h. We also aim to analyze the proposed incorporation of a decoder in SSL using other state-of-the-art SSL techniques like wavLM. Furthermore, we would like to evaluate the proposed model for other downstream tasks like phoneme recognition.

\section{Acknowledgement}
\label{ack}
We would like to thank Metilda N J for the technical discussion and her help in preparing this paper.

\bibliographystyle{IEEEtran}

\bibliography{mybib}

\begin{thebibliography}{10}
\providecommand{\url}[1]{#1}
\csname url@samestyle\endcsname
\providecommand{\newblock}{\relax}
\providecommand{\bibinfo}[2]{#2}
\providecommand{\BIBentrySTDinterwordspacing}{\spaceskip=0pt\relax}
\providecommand{\BIBentryALTinterwordstretchfactor}{4}
\providecommand{\BIBentryALTinterwordspacing}{\spaceskip=\fontdimen2\font plus
\BIBentryALTinterwordstretchfactor\fontdimen3\font minus
  \fontdimen4\font\relax}
\providecommand{\BIBforeignlanguage}[2]{{%
\expandafter\ifx\csname l@#1\endcsname\relax
\typeout{** WARNING: IEEEtran.bst: No hyphenation pattern has been}%
\typeout{** loaded for the language `#1'. Using the pattern for}%
\typeout{** the default language instead.}%
\else
\language=\csname l@#1\endcsname
\fi
#2}}
\providecommand{\BIBdecl}{\relax}
\BIBdecl

\bibitem{oord2018representation}
A.~V.~D. Oord, Y.~Li, and O.~Vinyals, ``Representation learning with
  contrastive predictive coding,'' \emph{arXiv preprint arXiv:1807.03748},
  2018.

\bibitem{chung2019unsupervised}
Y.-A. Chung, W.-N. Hsu, H.~Tang, and J.~Glass, ``An unsupervised autoregressive
  model for speech representation learning,'' \emph{arXiv preprint
  arXiv:1904.03240}, 2019.

\bibitem{schneider2019wav2vec}
S.~Schneider, A.~Baevski, R.~Collobert, and M.~Auli, ``wav2vec: Unsupervised
  pre-training for speech recognition,'' \emph{arXiv preprint
  arXiv:1904.05862}, 2019.

\bibitem{baevski2019vq}
A.~Baevski, S.~Schneider, and M.~Auli, ``vq-wav2vec: Self-supervised learning
  of discrete speech representations,'' \emph{arXiv preprint arXiv:1910.05453},
  2019.

\bibitem{liu2020mockingjay}
A.~T. Liu, S.-w. Yang, P.-H. Chi, P.-c. Hsu, and H.-y. Lee, ``Mockingjay:
  Unsupervised speech representation learning with deep bidirectional
  transformer encoders,'' in \emph{IEEE International Conference on Acoustics,
  Speech and Signal Processing (ICASSP)}.\hskip 1em plus 0.5em minus
  0.4em\relax IEEE, 2020, pp. 6419--6423.

\bibitem{baevski2020wav2vec}
A.~Baevski, Y.~Zhou, A.~Mohamed, and M.~Auli, ``wav2vec 2.0: A framework for
  self-supervised learning of speech representations,'' \emph{Advances in
  Neural Information Processing Systems}, vol.~33, pp. 12\,449--12\,460, 2020.

\bibitem{9053569}
M.~Ravanelli, J.~Zhong, S.~Pascual, P.~Swietojanski, J.~Monteiro, J.~Trmal, and
  Y.~Bengio, ``Multi-task self-supervised learning for robust speech
  recognition,'' in \emph{IEEE International Conference on Acoustics, Speech
  and Signal Processing (ICASSP)}, 2020, pp. 6989--6993.

\bibitem{9478264}
A.~T. Liu, S.-W. Li, and H.-y. Lee, ``Tera: Self-supervised learning of
  transformer encoder representation for speech,'' \emph{IEEE/ACM Transactions
  on Audio, Speech, and Language Processing}, vol.~29, pp. 2351--2366, 2021.

\bibitem{hsu2021hubert}
W.-N. Hsu, B.~Bolte, Y.-H.~H. Tsai, K.~Lakhotia, R.~Salakhutdinov, and
  A.~Mohamed, ``Hubert: Self-supervised speech representation learning by
  masked prediction of hidden units,'' \emph{IEEE/ACM Transactions on Audio,
  Speech, and Language Processing}, vol.~29, pp. 3451--3460, 2021.

\bibitem{chen2021wavlm}
S.~Chen, C.~Wang, Z.~Chen, Y.~Wu, S.~Liu, Z.~Chen, J.~Li, N.~Kanda,
  T.~Yoshioka, X.~Xiao \emph{et~al.}, ``Wavlm: Large-scale self-supervised
  pre-training for full stack speech processing,'' \emph{arXiv preprint
  arXiv:2110.13900}, 2021.

\bibitem{sennrich2015neural}
R.~Sennrich, B.~Haddow, and A.~Birch, ``Neural machine translation of rare
  words with subword units,'' \emph{arXiv preprint arXiv:1508.07909}, 2015.

\bibitem{lu2015study}
L.~Lu, X.~Zhang, K.~Cho, and S.~Renals, ``A study of the recurrent neural
  network encoder-decoder for large vocabulary speech recognition,'' in
  \emph{Sixteenth Annual Conference of the International Speech Communication
  Association}, 2015.

\bibitem{8462506}
L.~Dong, S.~Xu, and B.~Xu, ``Speech-transformer: A no-recurrence
  sequence-to-sequence model for speech recognition,'' in \emph{2018 IEEE
  International Conference on Acoustics, Speech and Signal Processing
  (ICASSP)}, 2018, pp. 5884--5888.

\bibitem{vaswani2017attention}
A.~Vaswani, N.~Shazeer, N.~Parmar, J.~Uszkoreit, L.~Jones, A.~N. Gomez,
  {\L}.~Kaiser, and I.~Polosukhin, ``Attention is all you need,''
  \emph{Advances in neural information processing systems}, vol.~30, 2017.

\bibitem{7953075}
S.~Kim, T.~Hori, and S.~Watanabe, ``Joint ctc-attention based end-to-end speech
  recognition using multi-task learning,'' in \emph{IEEE International
  Conference on Acoustics, Speech and Signal Processing (ICASSP)}, 2017, pp.
  4835--4839.

\bibitem{ott2019fairseq}
M.~Ott, S.~Edunov, A.~Baevski, A.~Fan, S.~Gross, N.~Ng, D.~Grangier, and
  M.~Auli, ``fairseq: A fast, extensible toolkit for sequence modeling,'' in
  \emph{Proceedings of NAACL-HLT 2019: Demonstrations}, 2019.

\bibitem{watanabe2018espnet}
S.~Watanabe, T.~Hori, S.~Karita, T.~Hayashi, J.~Nishitoba, Y.~Unno, N.~{Enrique
  Yalta Soplin}, J.~Heymann, M.~Wiesner, N.~Chen, A.~Renduchintala, and
  T.~Ochiai, ``{ESPnet}: End-to-end speech processing toolkit,'' in
  \emph{Proceedings of Interspeech}, 2018, pp. 2207--2211.

\bibitem{panayotov2015librispeech}
V.~Panayotov, G.~Chen, D.~Povey, and S.~Khudanpur, ``Librispeech: an asr corpus
  based on public domain audio books,'' in \emph{2015 IEEE international
  conference on acoustics, speech and signal processing (ICASSP)}.\hskip 1em
  plus 0.5em minus 0.4em\relax IEEE, 2015, pp. 5206--5210.

\end{thebibliography}


\end{document}